\documentclass[a4paper]{article}

\usepackage{INTERSPEECH2021}
\usepackage{url}
\usepackage{multirow}
\usepackage{subfigure}
\usepackage{graphicx}

\title{EMOVIE: A Mandarin Emotion Speech Dataset with a Simple Emotional Text-to-Speech Model}
\name{Chenye Cui$^1$, Yi Ren$^1$, Jinglin Liu$^1$, Feiyang Chen$^1$, Rongjie Huang$^1$,\\ Ming Lei$^2$, Zhou Zhao$^1$}
\address{
  $^1$Zhejiang University\\
  $^2$Alibaba Group}
\email{\{chenyecui,rayeren,jinglinliu,chenfeiyang,rongjiehuang\}@zju.edu.cn,\\ lm86501@alibaba-inc.com, zhaozhou@zju.edu.cn}

\begin{document}

\maketitle
\begin{abstract}
Recently, there has been an increasing interest in neural speech synthesis. While the deep neural network achieves the state-of-the-art result in text-to-speech (TTS) tasks, how to generate a more emotional and more expressive speech is becoming a new challenge to researchers due to the scarcity of high-quality emotion speech dataset and the lack of advanced emotional TTS model.  
In this paper, we first briefly introduce and publicly release a Mandarin emotion speech dataset including 9,724 samples with audio files and its emotion human-labeled annotation.   
After that, we propose a simple but efficient architecture for emotional speech synthesis called EMSpeech. Unlike those models which need additional reference audio as input, our model could predict emotion labels just from the input text and generate more expressive speech conditioned on the emotion embedding.
In the experiment phase, we first validate the effectiveness of our dataset by an emotion classification task. Then we train our model on the proposed dataset and conduct a series of subjective evaluations. Finally, by showing a comparable performance in the emotional speech synthesis task, we successfully demonstrate the ability of the proposed model.

\end{abstract}

\section{Introduction}
In the last few years, with the rapid development of text-to-speech (TTS) systems \cite{shen2018natural,ren2020fastspeech,chen2020multispeech,ren2019almost,xu2020lrspeech}, there has been a surge of interest in generating speeches with more emotional information. The specific objective of emotional TTS is to synthesize a more affectionate human-like speech or generate a speech audio sample with expected emotions. 
And with the ability to generate expressive speeches, the emotional TTS system can be applied to some applications, such as movie dubbing, that could only be done by people with specific skills.

At present, many TTS datasets like VCTK \cite{veaux2016superseded}, LJSpeech \cite{ljspeech17} and VCC2018 \cite{lorenzo2018voice} have been released and have made a significant contribution to speech synthesis tasks. And for Chinese Mandarin tasks, CSMSC\footnote{https://www.data-baker.com/open\%20source.html\#/data/index/source}, AISHELL-3 \cite{shi2020aishell} and DiDiSpeech \cite{guo2020didispeech} are published in recent years.
Some audio-visual datasets like EmoTV1 \cite{abrilian2005emotv1}, ENTERFACE \cite{martin2006enterface}, HUMAINE \cite{douglas2007humaine}, RML \cite{wang2008recognizing}, VAM \cite{grimm2008vera}, CREMA-D \cite{cao2014crema}, RAVDESS \cite{livingstone2018ryerson} and MELD \cite{poria2018meld} have been established for the multi-model emotional tasks at an earlier stage.

\begin{table}[t]
  \caption{Publicly available emotional speech datasets}
   \vspace{-2mm}
  \label{tab:datasets}
  \centering
  \small
  \begin{tabular}{c c c c}
    
    \toprule
    \multicolumn{1}{c}{\textbf{Name}} &  
    \multicolumn{1}{c}{\textbf{Language}} &
    \multicolumn{1}{c}{\textbf{Hours}} &
    \multicolumn{1}{c}{\textbf{Sampling Rate}} \\
    \midrule
    \textit{AIBO}&\textit{Multi}&\textit{9.33}&\textit{44.1 kHz}\\
    \textit{PRIORI}&\textit{English}&\textit{25.20}&\textit{8 kHz}\\
    \textit{ESD}&\textit{Multi}&\textit{0.28}&\textit{16 kHz}\\
    \textit{LSSED}&\textit{English}&\textit{206.42}&\textit{16 kHz}\\
    \textit{Emov-DB}&\textit{Multi}&\textit{Not Given}&\textit{16 kHz}\\
    \midrule
    \textit{EMOVIE}&\textit{Mandarin}&\textit{4.18}&\textit{22.05 kHz}\\
    \bottomrule
  \end{tabular}
   \vspace{-5mm}
\end{table}

As shown in Table~\ref{tab:datasets}, several datasets have been released to conduct emotion-related speech tasks since 2004.
AIBO \cite{batliner2004you} is a cross-linguistic corpus collecting from the interaction between children and robots.
PRIORI \cite{khorram2018priori} dataset is recorded from natural conversations during daily smartphone usage and annotates the emotion activation and emotion valence using a 9-point Likert scale.
ESD \cite{zhou2020seen} is the first parallel multi-lingual and multi-speaker emotional speech dataset designed for voice conversion tasks and contains five emotional classes in each language.
LSSED \cite{fan2021lssed} is a large English dataset designed for SER tasks, which have a nine-classes emotion annotation.
EmoV-DB \cite{adigwe2018emotional} is the first dataset designed for emotional TTS tasks. It is acted by five speakers in two different languages and annotates the audio in five classes of emotions.
Among them, no one is collected from the broadcasting television programs or movies, AIBO and PRIORI are collected in a natural environment, and others are acted by specific speakers. Unfortunately, there is no public dataset designed for Mandarin emotional TTS tasks.

In earlier researches, some deep neural methods like Tacotron \cite{shen2018natural,wang2017tacotron}, FastSpeech \cite{ren2020fastspeech,ren2019fastspeech}, and Waveglow \cite{prenger2019waveglow} have shown their ability to generate fluent speech audio samples as well as expressive singing voice~\cite{ren2020deepsinger,liu2021diffsinger} in real-time.
Recently, several works have been published to generate emotional speech audio samples and already achieve some signs of progress.
The most common methods use emotion embedding, which is extracted with the assistance of a Global Style Token \cite{wang2018style} architecture or a speech emotion recognition (SER) model to get a better performance in generating expressive speech.
Li's study \cite{li2021controllable} uses a reference encoder and an auxiliary network to generate the emotional embedding, and calculates a style loss between the predicted spectrograms and the target spectrograms.
Wu's work \cite{wu2019end} uses the weight of the attention layer in the GST as the probability of each emotion label to calculate the cross-entropy loss and generate the emotion token.
Some studies \cite{cai2020emotion} use the speech emotion recognition module as an alternative approach to modeling the emotion feature from existing audio.
Several other studies \cite{aggarwal2020using,zhang2020adadurian} use VAE or DurIAN to model the emotional information and have achieved some progress.
The idea most similar to ours is TP-GST \cite{stanton2018predicting} and FET \cite{lei2020fine} which also predict emotion embedding from the text, but they use extra audio as a reference input when training. In contrast, our model predicts the emotion information from the text context.

However, there are also two remarkable problems in current studies: 1) Many of the emotion speech datasets are designed for speech emotion recognition tasks, which always have ambiguous, unclear, or incomplete speech audio samples and hard to use in speech synthesis. And the high-quality speech datasets for TTS tasks always do not have emotional annotations and cannot be used in emotion tasks. 2) Almost all of the current emotional speech synthesis works need additional reference audio or emotional label input to generate expressive speeches. Thus, the input sequence is often given an inappropriate emotion which always leads to a strange synthesis result.


To solve the first problem, we collect and annotate a new Mandarin emotion speech dataset. The dataset is collected from some movies with natural and expressive speeches and it has an equivalent audio quality to some recording datasets with more natural emotion expression. To our best knowledge, the dataset is the first public Mandarin speech dataset designed for the emotional TTS tasks, and it also can be used for several other emotion-related speech tasks such as speech emotion transfer or SER. The dataset is publicly available at our Github page\footnote{https://viem-ccy.github.io/EMOVIE/dataset\_release}.

To manage the second issue, we propose EMSpeech, an experimental architecture based on the FastSpeech 2 \cite {ren2020fastspeech}, which models a latent mapping from text content to the emotion of the speech, including the duration and the pitch of each phoneme. At the same time, the proposed model could automatically predict emotion through the text input during inference and generates a natural and expressive speech audio sample. Besides, our architecture supports controllable emotional TTS which can receive a manually given emotion label input to generate a speech audio sample with the expected emotion.

In the experiment, we first validate the availability of our dataset by testing the classification accuracy. After that, we make a series of comparative experiments to verify the effectiveness of the proposed method. In order to intuitively see the result of the emotion predicting and emotion controlling, we also analyze the mel-spectrogram and the pitch of the generated speech audio samples.

The main contributions of this paper include: 1) We briefly introduce and publicly release a speech dataset with emotional annotations, which is the first Mandarin movie speech dataset designed for emotional TTS tasks. 2) We establish a simple model for emotional speech synthesis, which achieves a good performance at emotion expression of generated speech.

\section{EMOVIE Dataset}
In this section, we will introduce EMOVIE, a new Mandarin emotion speech dataset that has high-quality natural emotion speech data and suitable for emotional text-to-speech and some other tasks.  
\subsection{Data Obtaining and Processing}
We get these speech samples from seven Mandarin movies that have a relatively noise-free audio track. To our best knowledge, compared with action movies and war movies, feature movies and comedy movies always have a clearer dialogue environment, so most of the movies are chosen from these kinds of movies. The raw audio with 5.1 channel or 7.1 channel audio tracks is extracted from the movie files using the \textit{ffmpeg} tool. And then we execute the audio track decomposition to get the \textit{Front Center} channel which always has a higher quality dialog audio with less environmental noise.

As we know, embedded subtitles or third-part subtitle files always include the timestamp of each line in the movie. This characteristic can help us cut the full audio track to a single-sentence speech audio sample and correspond the text to its audio fragment.

Moreover, we have further screened the speech audio samples very carefully. At last, we get 9724 samples, 4.18 hours of audio in total.
\subsection{Data Annotation}

\begin{table}[th]
   \vspace{-4mm}
  \caption{Polarity of the emotion in the dataset}
   \vspace{-2mm}
  \label{tab:polarity}
  \centering
  
  \begin{tabular}{c c}
    \toprule
    \multicolumn{1}{c}{\textbf{Polarity}} &  
    \multicolumn{1}{c}{\textbf{Meaning}}  \\
    \midrule
    \multirow{1}*{-1} & The audio has an absolute negative emotion.\\
    \midrule
    \multirow{1}*{-0.5} & The audio has an slight negative emotion.\\
    \midrule
    \multirow{1}*{0} & The audio has an neutral emotion.\\
    \midrule
    \multirow{1}*{0.5} & The audio has an slight positive emotion.\\
    \midrule
    \multirow{1}*{1} & The audio has an absolute positive emotion.\\
    \bottomrule
  \end{tabular}
   \vspace{-2mm}
\end{table}


We use -1, -0.5, 0, 0.5 and 1 to annotate the emotion polarity of the speech audio sample, and the description of the polarity of the emotion in the annotation is shown in Table~\ref{tab:polarity}.

To keep the annotation clearer, the annotators are not allowed to see the text content of the corresponding audio file, which keeps the annotations only obtained by the inherent factors of the audio.

\subsection{Data Analysis}

\vspace{-2mm}
\begin{figure}[htbp]
    \centering
    \vspace{-4mm}
    \subfigure[Polarities distribution]{
        \begin{minipage}[t]{0.43\linewidth}
        \centering
        \includegraphics[width=\linewidth]{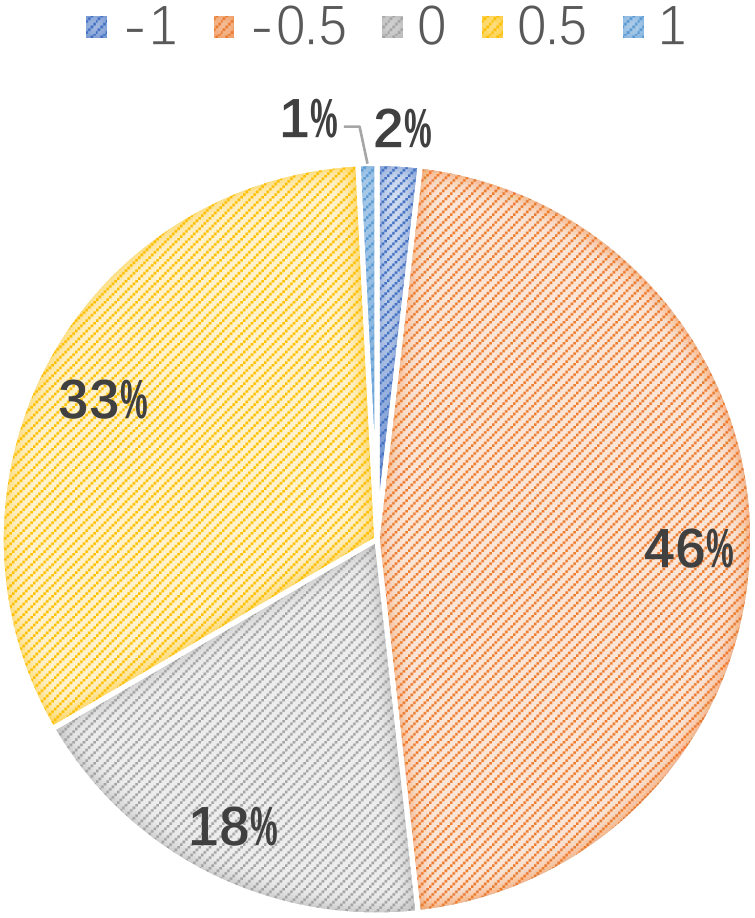}
        \label{fig:emotion}
        \end{minipage}%
    }%
    \subfigure[Length distribution]{
        \begin{minipage}[t]{0.57\linewidth}
        \centering
        \includegraphics[width=\linewidth]{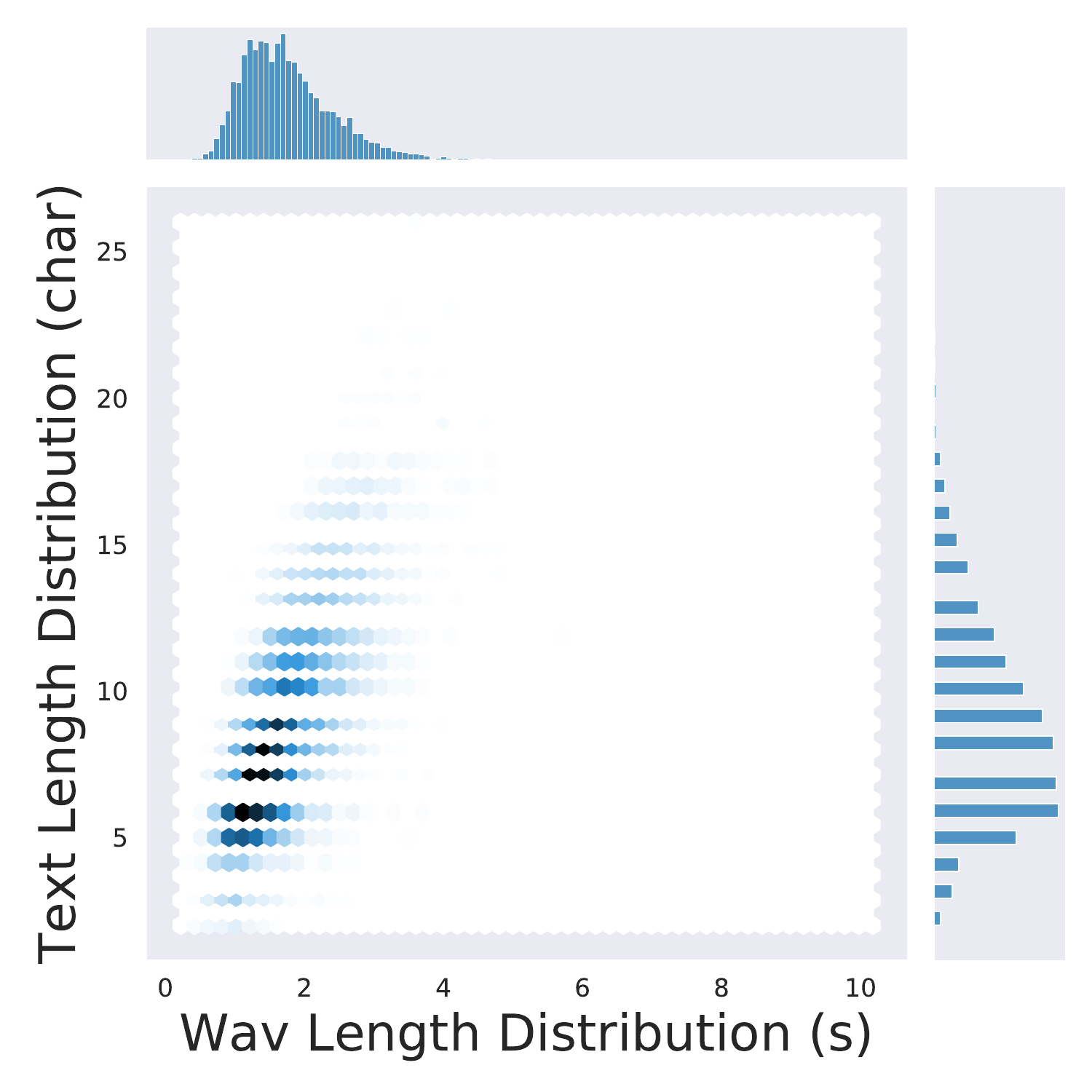}
        \label{fig:length}
        \end{minipage}%
    }%
    \vspace{-3mm}
    \caption{Distributions of the dataset}
    \vspace{-3mm}
\end{figure}

As shown in Fig.~\ref{fig:emotion}, we give the emotion polarities distribution of our dataset. 
Samples with the polarity of `-0.5' and `0.5' are accounted for a larger proportion, which contains 4573 and 3171 samples respectively. The samples of these 2 polarities account for 79\% of the total samples. The label second only to them is `0', which has 1783 samples. And there are only 179 and 78 samples for the very obvious polarity `-1' and `1'.

We also plot the joint distribution of the audio samples length and the text length in Fig.~\ref{fig:length}. And the average length of the audio samples and text is 1.78 seconds and 8.93 characters respectively.

\section{Model Architecture}
\begin{figure*}[htbp]
    \centering
    \vspace{-2mm}
    \subfigure[Pipeline of proposed model] {
        \begin{minipage}[c]{0.38\linewidth}
        \centering
        \includegraphics[width=\linewidth]{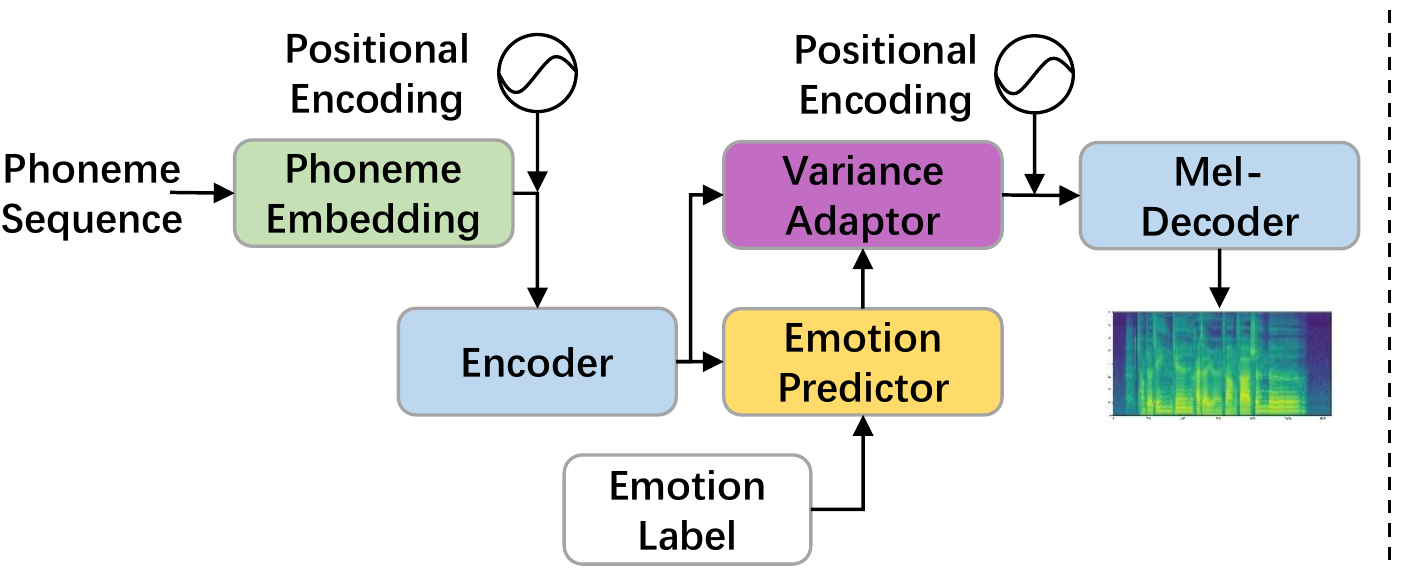}
        \label{fig:model}
        \end{minipage}%
    }%
    \subfigure[Emotion Predictor architecture] {
        \begin{minipage}[c]{0.62\linewidth}
        \centering
        \includegraphics[width=\linewidth]{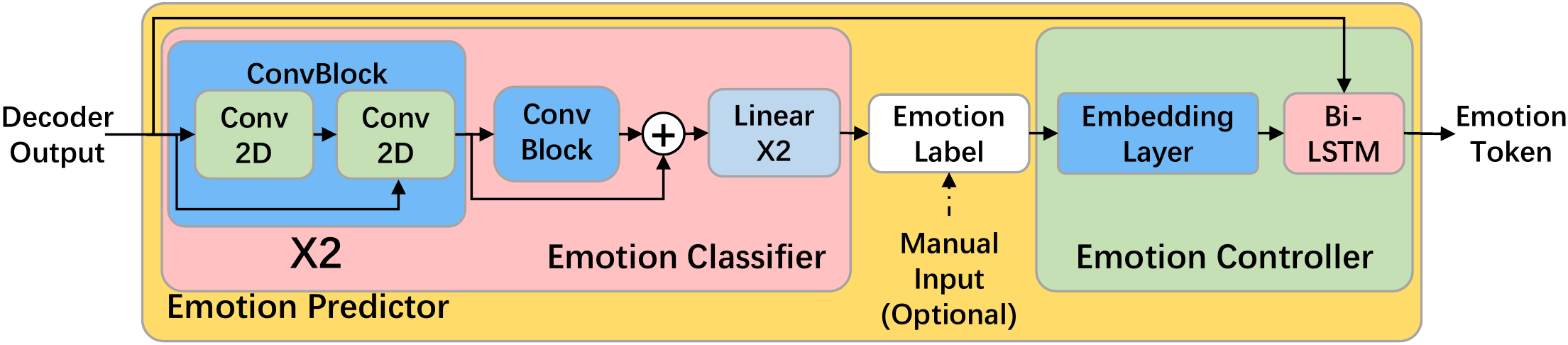}
        \label{fig:predictor}
        \end{minipage}%
    }%
    \vspace{-4mm}
    \caption{Model architecture}
    \vspace{-4mm}
\end{figure*}

Fig.~\ref{fig:model} shows the framework of our proposed EMSpeech for training and inference. It shares a similar architecture with the FastSpeech 2 \cite{ren2020fastspeech} which is a fast and high-quality end-to-end TTS system.

In this paper, since the emotional annotations in our dataset do not directly contain any information from the text, we regard the polarity of emotion as an attribute of speech and try to model the implicit mapping between text information and latent speech emotion representations. Finally, we use the emotion information to control the prosody including duration and the pitch, and then emotional speech samples can be generated. 

In detail, we use an emotion predictor on the proposed model, which could predict the emotion of the input text and could encode the emotion label to a latent emotion embedding. By adding the emotion embedding to the output of the encoder, the duration predictor and pitch predictor could make the speech sample have a better emotion expression.

The emotion predictor contains two main components: 1) an emotion classifier and 2) an emotion controller. 

Also, the emotion controller can receive either the predicted emotion label or the external specific emotion label as input at inference stage, which means the model not only can generate an expressive speech audio sample with an automatically predicted emotion but also can generate a speech audio sample with an expected emotion.

Besides, a speaker embedding is added after the duration predictor, which is adapted for multi-speaker generating.

\subsection{FastSpeech 2}
The components of the FastSpeech 2 include: a phoneme embedding module, a encoder, a variance adaptor and a mel-spectrogram decoder.
The encoder converts the phoneme embedding sequence into the phoneme hidden sequence. The variance adaptor adds different variance information such as duration, pitch and energy into the hidden sequence. The mel-spectrogram decoder converts the adapted hidden sequence into mel-spectrogram sequence in parallel.

\subsection{Emotion Classifier}

The effectiveness of the emotion predictor mainly depends on the emotion classifier. And we use the ResNet-like network, which is simple enough and has been proven its ability on classification tasks, to predict the emotion label from the input text.

As shown in Fig.~\ref{fig:predictor}, the framework of the emotion classifier is a 3-layer ConvBlock stack with skip connect operations among the layers. Each ConvBlock has double 2D convolution layers followed by a 2D pooling operation.

\subsection{Emotion Controller}
The emotion controller aims to map a specific emotion label to an emotion embedding. Considering the emotion embedding is related to a specific phoneme and its position of the input sequence, we choose a network with recurrent structures to encode a sequence-length emotion embedding.

As shown in Fig.~\ref{fig:predictor}, the Emotion controller uses an Embedding network to model an intermediate feature vector used as the initial hidden state of a Bi-LSTM network. Then we use the Bi-LSTM network to model the phoneme-level emotion embedding for the whole input sequence. 

\section{Experiment}
In this section, we give the details of our experiments on the proposed dataset. 
Although visualizations for some experiments result is provided below, we recommend readers to listen to the audio samples published on the demo page\footnote{https://viem-ccy.github.io/EMOVIE/} to have a more intuitive impression.

\subsection{Experimental Setup}

\subsubsection{Dataset Preparation} 
We first randomly spilt our dataset into 3 sets: 9,524 samples for training, 100 samples for validation and 100 samples for testing. 
For the audio data, we convert the audio waveform into 80-dimensional mel-spectrograms for the emotional text-to-speech model.
The frame size and hop size of the mel-spectrograms are set to 1024 and 256 respectively.
Following the approach in previous works, we use an open-source grapheme-to-phoneme tool\footnote{https://github.com/Kyubyong/g2p} to convert the text sequence into the phoneme sequence.
After that, we use \textit{Montreal Forced Aligner}(MFA) \cite{mcauliffe2017montreal} tool to extract the phoneme duration.
The emotion data are also processed during the data preparation. We choose the emotion polarity as the emotion information input and convert them to real number labels.

\subsubsection{Model Configuration}
Our emotional TTS model is based on FastSpeech 2 \cite {ren2020fastspeech} which consists of 4 feed-forward Transformer (FFT) blocks in the encoder and the mel-spectrogram decoder. 
Following FastSpeech 2, for each FFT block, both the dimension of phoneme embedding and the hidden size of the self-attention are set to 256. 
The number of attention heads is set to 2 and the kernel sizes of the 1D-convolution in the 2-layer convolutional network after the self-attention layer are set to 9 and 1. The input/output sizes of the first layer and the second layer in the 2-layer convolutional network are set to 256/1024 and 1024/256 respectively.
For the variance predictor, the kernel sizes of the 1D-convolution are set to 3, with input/output sizes of 256/256 for both layers and the dropout rate is set to 0.5. 

In the emotion classifier, all the kernel sizes of the 2D-convolution are set to 3x3 and both of the stride and padding sizes are set to 1. And the input/output dimensions of the 3 ConvBlocks are 1/64, 64/128, 128/256 respectively. 
In the emotion controller, the dimension of the Bi-LSTM layer's input is set to 512 which equal to the size of the encoder layer output plus the size of Embedding layer output. And the size of the hidden state is set to 256. 

\subsubsection{Training and Inferring}
Both the baseline FastSpeech 2 model and our proposed model are trained on a single NVIDIA RTX2080Ti GPU with a batch size of 16 sentences. Adam optimizer is applied to our model and the configuration is set to $\beta_{1}=0.9$,$\beta_{2}=0.98$ and $\epsilon=10^{-9}$, and the learning rate schedule is set as same as the FastSpeech 2 configuration. Besides, the classifier is optimized with cross-entropy loss, the variance adaptor is optimized with mean square error (MSE) loss and other parts of the model are optimized with mean absolute error (MAE) loss.

To get a more precise emotion modeling, the emotion controller uses a true emotion label as conditional inputs to model the emotion embedding instead of using the predicted label when training. In inference, we use values predicted by the classifier that are jointly trained with the EMSpeech model.

During the inference, the output 80-dimensional mel-spectrograms of our model are transformed into waveform audio by a pre-trained Parallel WaveGAN \cite{yamamoto2020parallel}.

\subsection{Experiment Result}
\subsubsection{Emotion Classification on EMOVIE}
To verify the reliability of the annotation of the dataset, we conduct an experiment about classification accuracy. We use the output emotion classifier which is jointly trained with the model to calculate the accuracy.
The network exhibits a comparable result of emotion polarity classification which achieves an accuracy rate up to 48.2\%. Since the accuracy is significantly higher than the random classification, we show that the annotations of the dataset is available.


\subsubsection{Evaluation on Audio with Predicted Emotion}
\begin{figure}[h]
  \centering
   \vspace{-4mm}
  \includegraphics[width=\linewidth]{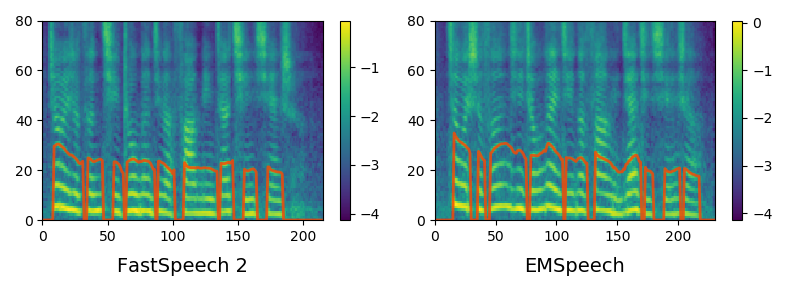}
  \vspace{-7mm}
  \caption{Generating result of emotion predicting}
  \vspace{-3mm}
  \label{fig:emo}
\end{figure}

To prove that our model can generate more emotional speech audio sample with the assistance of the emotion predictor, we conduct a experiment between the FastSpeech 2 and our EMSpeech.
In this experiment, we generate speech audio samples using FastSpeech 2 and our model with the same text input, and next evaluate their emotional expression.
We visualize the comparison of spectrograms in Fig.~\ref{fig:emo}. We can see the generating result of EMSpeech has a greater pitch fluctuation than that result from the FastSpeech 2. 
The scale of the MOS is set between 1 to 5 in emotion predicting results evaluation. MOS from 1 to 5 denotes No, Slight, Ordinary, Good, and Strong expressiveness respectively in the evaluation on audio with predicted emotion.
And the MOS in Table~\ref{tab:MOS2} illustrates our model can generate speech audio samples containing more emotional expressions.
We show that our EMSpeech has the ability to model the mapping between the input text and the speech emotion, which yields more expressive speech than a FastSpeech 2 conditioned on the same data.

\begin{table}[h]
    \vspace{-1mm}
    \caption{MOS of emotion predicting result}
    \label{tab:MOS2}
    \vspace{-2mm}
    \centering
    \begin{tabular}{c|c}
    \toprule  
    \textbf{FastSpeech 2} & \textbf{EMSpeech}\\
    \midrule  
    $3.45$&$3.88$\\
    \bottomrule 
    \end{tabular}
    \vspace{-5mm}
\end{table}

\subsubsection{Analyses on Emotion Controllability}

\begin{figure}[h]
  \centering
   \vspace{-4mm}
  \includegraphics[width=\linewidth]{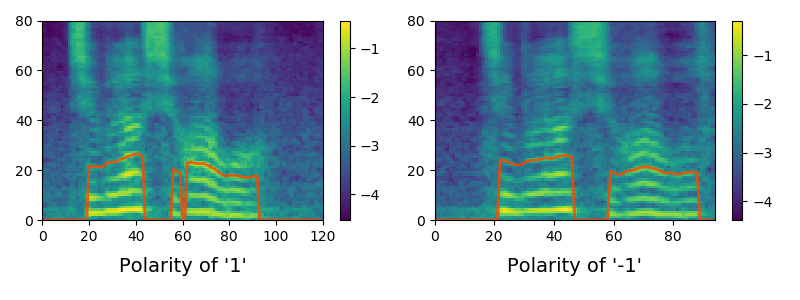}
   \vspace{-7mm}
  \caption{Generating result of emotion controlling}
   \vspace{-3mm}
  \label{fig:pol}
\end{figure}

We conduct a emotion controllability experiment to further prove that the label we manually input can control the emotion of the generated audio.
In the emotion controlling experiment, we use two different emotion labels `1' and `-1' to generate speech audio samples with the same text content and compare their emotional tendencies to verify.
Fig.~\ref{fig:pol} gives a comparison of spectrograms and pitch between the audio sample generated with the two labels. From the figure we can see the audio with the polarity of `1' often has a shorter phoneme duration with a rising pitch trend compared to the audio with the polarity of `-1'. 
For subject evaluation, we choose the mean opinion score (MOS) to evaluate polarity of the synthesized speeches by our emotional TTS model. 
In emotion controllability evaluation the scale of MOS is set to between -1 to 1, where -1 denotes strong negative and 1 denotes strong positive polarity. 
And as shown in Table~\ref{tab:MOS1}, the MOS shows that the samples generated with the polarity of `-1' often have a more negative emotion than the samples generated with the polarity of `1', indicating that our method can generate controllable speech in excepted polarities.

\begin{table}[h]
    \vspace{-1mm}
    \caption{MOS of emotion controlling result}
    \label{tab:MOS1}
    \vspace{-2mm}
    \centering
    \begin{tabular}{c|c}
    \toprule  
    \textbf{Polarity label of `1'}  & \textbf{Polarity label of `-1'}\\
    \midrule  
    $0.28$&$-0.40$\\
    \bottomrule 
    \end{tabular}
    \vspace{-5mm}
\end{table}

\section{Conclusion}
In this paper, we introduced and publicly released a new Mandarin movie speech dataset with emotion polarity annotations.
Moreover, an emotional text-to-speech system based on FastSpeech 2 was proposed and achieved a good performance. According to the experimental result, we successfully demonstrated the availability of our EMOVIE dataset and the effectiveness of our proposed EMSpeech in terms of generating speech with more emotion expression and controlling the emotion of generated speech. 
Even though the system of EMSpeech is a very simple one, the result shows the efficiency of the EMOVIE dataset which is promising for future work. In the future, we will enhance our dataset and improve our model, and seek new breakthroughs in emotion-related tasks.

\bibliographystyle{IEEEtran}

\bibliography{mybib}

\end{document}